\documentclass[10pt,twocolumn,letterpaper]{article}

\usepackage{cvpr}
\usepackage{times}
\usepackage{epsfig}
\usepackage{graphicx}
\usepackage{amsmath}
\usepackage{amssymb}
\usepackage{overpic}
\usepackage{subfigure}
\usepackage{multirow}

\newcommand{\C}[1]{{\color{black}#1}}

\usepackage[pagebackref=true,breaklinks=true,letterpaper=true,colorlinks,bookmarks=false]{hyperref}

\cvprfinalcopy 

\def\cvprPaperID{492} 

\begin{document}

\title{Deep Joint Rain Detection and Removal from a Single Image}

\author{Wenhan~Yang,
	Robby T. Tan, 
	Jiashi~Feng, 
	Jiaying~Liu, 
	Zongming~Guo, 
	and~Shuicheng~Yan 
}

\maketitle

\begin{abstract}
In this paper, we address a rain removal problem from a single image,
even in the presence of heavy rain and rain streak accumulation.
Our core ideas lie in the new rain image models and a
novel deep learning architecture. 
We first modify an existing model comprising a rain streak layer and a background layer, by adding a binary map that locates rain streak regions.
Second, we create a new model consisting of a component
representing rain streak accumulation (where individual streaks cannot be
seen, and thus visually similar to mist or fog), and another component
representing various shapes and directions of overlapping rain streaks,
which usually happen in heavy rain.  
Based on the first model, we develop a multi-task deep learning
architecture that learns the binary rain streak map, the appearance of
rain streaks, and the clean background, which is our ultimate
output. The additional binary map is critically beneficial, since its
loss function can provide additional strong information to the
network. To handle rain streak accumulation (again, a phenomenon
visually similar to mist or fog) and various shapes and directions of
overlapping rain streaks, we propose a recurrent rain detection and removal
 network that removes rain streaks and clears up the
rain accumulation iteratively and progressively. 
In each recurrence of our method, a new contextualized dilated 
network is developed to exploit regional contextual information
and outputs better representation for rain detection.
The evaluation on real images, particularly on heavy rain,
shows the effectiveness of our novel models and architecture,
outperforming the state-of-the-art methods significantly. Our codes and data sets will be publicly available.
\end{abstract}

\section{Introduction}\label{sec:introduction}
Restoring rain images is important for many computer vision applications in outdoor scenes. 
Rain degrades visibility significantly and causes many computer vision systems to likely fail.  Generally,
rain introduces a few types of visibility degradation. Raindrops
obstruct, deform and/or blur the background scenes.
Distant rain streaks accumulate and generate atmospheric veiling
effects similar to mist or fog, which severely reduce the visibility
by scattering light out and into the line of sight.
Nearby rain streaks exhibit strong specular highlights that
occlude background scenes. These rain streaks can have various
shapes and directions, particularly in heavy rain, causing severe
visibility degradation.

\begin{figure}[t]
	\centering	
	\includegraphics[width=8.3cm]{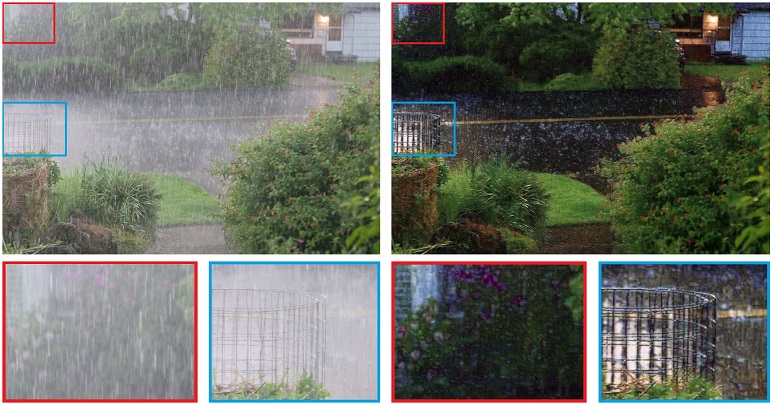}	
	\caption{An example result of our proposed rain removal method that removes heavy rain streaks and enhances the visibility significantly. Top: The raw image with rain streaks (left) and the output image of our method (right). Bottom: closer looks at specific regions (blue and red) for more details. }
	\label{fig:teaser}
	\vspace{-4mm}
	
\end{figure}

In the past decades, many researchers have devoted their attention to
solving the problem of restoring rain images. Some focus on rain image recovery from video sequences~\cite{barnum2010analysis,bossu2011rain,brewer2008using,chen2013generalized,Eigen_2013_ICCV,garg2004detection,garg2006photorealistic,garg2007vision,zhang2006rain}. Others focus on rain removal from the single image, by regarding the rain streak removal problem as a signal
separation problem ~\cite{ID,huang2014self,sun2014exploiting,chen2013generalized,luo2015removing}, or by relying on nonlocal mean smoothing~\cite{kim2013single}. 
While there are varying degrees of success, majority of existing
methods suffer from several limitations:
\begin{itemize}
	\item 
	Due to the intrinsic
	overlapping between rain streaks and background texture patterns, most
	methods tend to remove texture details in non-rain regions, leading to
	over-smoothing the regions.  
	\item
	The degradation of rain is complex, and the existing rain model widely
	used in previous methods~\cite{ID,chen2013generalized} is insufficient to cover some important
	factors in real rain images, such as the atmospheric veils due to rain streak
	accumulation, and different shapes or directions of streaks.
	\item
	The basic operation of many existing algorithms is on a local
	image patch or  a limited receptive field (a limited spatial
	range). Thus, spatial  contextual information in larger
	regions, which has been proven to be 
	useful for rain removal~\cite{2012_Huang_Context}, is rarely used.  
\end{itemize}  
Considering these limitations, our goal is to develop a novel rain model that is more capable of describing various rain
conditions in real scenes, including rain streak accumulation and heavy rain,
and then, use them to design an effective deep learning
architecture. Here, we focus on a single input image.  

To achieve the goal,  we explore the possible rain models and  deep
learning architectures that can effectively restore rain images even
in the presence of heavy rain. Our ideas are as follows.
First, we introduce novel region-dependent rain models.
In the models, we
use a rain-streak binary map, where 1 indicates the presence of
individually visible rain streaks in the pixels, and 0 otherwise. We
also model the appearance of rain streak accumulation (which is similar to
that of mist or fog), and the various shapes  and directions of
overlapping streaks, to simulate heavy rain. 

Second, based on our introduced models, we construct a deep network
that jointly detects and removes rain. Rain streak regions are
detected automatically and are used to constrain the rain
removal. With this, our network is capable of performing an adaptive
operation on rain and non-rain regions, preserving richer details.  

Third, to retrieve more contextual information, we propose a contextualized dilated network to
enlarge the receptive field. In this network, the
extracted features are enhanced in each recurrence progressively by
the aggregated information from several parallel dilated
convolutions.

Finally, to restore images captured in the environment
with both rain accumulation and various rain streak directions, we
propose a recurrent rain detection and removal network that
progressively removes rain streaks. Extensive experiments and
evaluations demonstrate that our method outperforms state-of-the-art
methods significantly on both synthesized data and real
data. Particularly for some heavy rain images, our method 
achieves considerably good results. 

Hence, our contributions are:
\begin{enumerate}
	\item
	The first method to model the rain-streak binary mask, and also to
	model the atmospheric veils due to rain streak accumulation as well as
	various shapes  and directions of overlapping rain
	streaks. This enables us to synthesize more similar data to real rain images for the network training.
	\item
	The first method to jointly detect and remove rains from single
	images.
	With the additional information of detected rain regions, our rain removal achieves better performance. 
	\item
	The first rain removal method that uses a contextualized dilated network to obtain more context while preserving rich local
	details. 
	\item
	The first method that addresses heavy rain by introducing an
	recurrent rain detection and removal network, where it removes rain
	progressively, enabling us to obtain good results even in
	significantly complex cases.
\end{enumerate}

Our training and testing data, as well as our codes, will be publicly available.


\section{Related Work}
\label{sec:related_work}

Compared with the video based deraining problem, the single image based problem is more ill-posed, due to the lack of temporal information.
Some single-image based rain removal methods regard the problem as a
layer separation problem.
Huang \emph{et al.}~\cite{ID} attempted to  separate the rain streaks
from the high frequency layer by sparse coding, with a learned
dictionary from the HOG features. However, the capacity of
morphological component analysis, the layer separation, 
and learned dictionary are limited. Thus, it usually causes the
over-smoothness of the background.
In~\cite{chen2013generalized}, a generalized low rank model is
proposed, where the rain streak layer is assumed to be low rank.
In~\cite{nonlocal_derain}, Kim~\emph{et al.} first detected rain streaks
and then removed them with the nonlocal mean filter.
In~\cite{luo2015removing}, Luo~\emph{et al.} proposed a
discriminative sparse coding method to separate rain streaks from
background images.
A recent work of~\cite{li2016rain} exploits the Gaussian 
mixture models to separate the rain streaks, achieving the
state-of-the-art performance, however, still with slightly smooth
background. 

In recent years, deep learning-based image processing applications
emerged with promising performance. These applications include
denoising~\cite{2010_Pascal_SDA,2012_Burger_MLPDenoising1,2012_Burger_MLPDenoising2,2009_Jain_ConvDenoise,2013_Forest_MCRobustDenoise},
completion~\cite{2012_Xie_NIPS},
super-resolution~\cite{2015_Dong_SRCNN,2014_Dong_SRCNN,2014_Cui_DNC,2014_Loo_Fast_Approx},
deblurring~\cite{2014_Schuler_Deblur},
deconvolution~\cite{2014_Li_Deconvolution} and style
transfer~\cite{2015_Gatys_NeuralArt,2015_Yan_PhotoAdjust},
\textit{etc}. There are also some recent works on bad weather
restoration or image enhancement, such as
dehazing~\cite{DehazeNet,multi_scale_dehaze}, rain drop and
dirt removal~\cite{Eigen_2013_ICCV} and light
enhancement~\cite{lore2015llnet}. Besides, with the superior modeling
capacity than shallow models, DL-based methods begin to solve
harder problems, such as blind image
denoising~\cite{2016arXiv160803981Z}. In this paper, we use deep
learning to jointly detect and remove rain.

\section{Region-Dependent Rain Image Model}
\label{sec:image_model}

We briefly review the commonly used rain
model, and generalize it to explicitly include a rain-streak binary 
map. Subsequently, we introduce a novel rain model that captures rain
accumulation (atmospheric veils) and rain streaks that have various shapes and directions, which are absent in the existing rain models.

\subsection{Region-Dependent Rain Image Formation}
\label{sec:mask_image_model}

The widely used rain model~\cite{li2016rain,luo2015removing,2012_Huang_Context} is expressed as:
\begin{equation}
	\label{eq:rain_gen1}
	\mathbf{O} = \mathbf{B} + \widetilde{\mathbf{S}},
\end{equation}
where $\mathbf{B}$ is the background scene without rain
streaks, and $\widetilde{\mathbf{S}}$ is the rain streak
layer. $\mathbf{O}$ is the captured image with rain streaks.
Based on this model, rain removal is regarded as a two-signal
separation problem. Namely, given the observation $\mathbf{O}$,
removing rain streaks is to estimate the background B and rain streak $\widetilde{\mathbf{S}}$,
based on the different characteristics of the rain-free images and rain
streaks. 
Existing rain removal methods relying on Eq.~\eqref{eq:rain_gen1} suffer from following two deficiencies. First,
$\widetilde{\mathbf{S}}$ has heterogeneous density, and heavy rain regions are much denser than light rain regions. 
Thus, it is hard to model $\widetilde{\mathbf{S}}$ with a uniform sparsity level assumption, which is needed for most of
existing sparsity-based methods.
Second, solving the signal separation
problem in Eq.~\eqref{eq:rain_gen1} without distinguishing the rain and non-rain regions will cause over-smoothness on the rain-free regions. The main reason for these difficulties lies in the intrinsic complexity to model~$\widetilde{\mathbf{S}}$. In Eq.~\eqref{eq:rain_gen1}, $\widetilde{\mathbf{S}}$ needs to model both location and intensity of rains, thus it is hard for the existing methods to jointly to localize and remove the rain streaks.

To overcome these drawbacks,  we first propose a generalized rain model
as follows: 
\begin{equation}
	\label{eq:rain_gen_two}
	\mathbf{O} = \mathbf{B} + \mathbf{SR},
\end{equation}
which includes a new region-dependent variable $\mathbf{R}$ to
indicate the location of individually 
visible rain streaks. Here elements in $\mathbf{R}$ takes binary values, where $1$
indicates rain regions and $0$ indicates non-rain
regions. Note that, although $\mathbf{R}$ can be easily estimated from $\mathbf{S}$ via hard-thresholding, modeling $\mathbf{R}$ separately from $\mathbf{S}$ provides two desirable benefits for learning based rain removal methods:
(1) it gives additional information for the network to learn about rain streak regions,
(2) it allows a new rain removal pipeline 
to detect rain regions first, and then to operate differently on rain-streak and non-rain-streak regions, preserving background details.

\subsection{Rain Accumulation and Heavy Rain}
\label{sec:extend_mask_image_model}

\begin{figure}[t]
	\subfigure[Heavy rain. ]{	
		\centering\includegraphics[height=2.2cm]{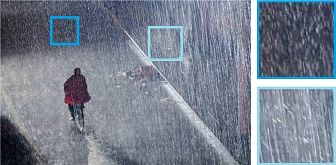}}
	\subfigure[Rain acccumulation. ]{	
		\centering\includegraphics[height=2.2cm]{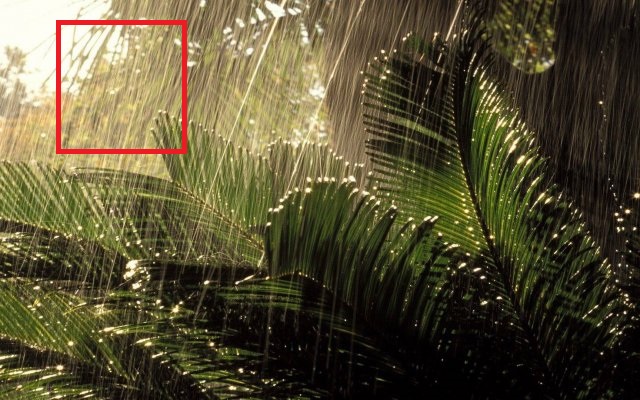}}
	\caption{(a) In heavy rain cases, the rain streaks have various shapes and directions~(shown in blue windows). (b) Rain accumulation reduces the visibility for distant scenes~(shown in red windows).}
	\vspace{-5mm}
	\label{fig:real_rain_cases}
\end{figure}

The rain image model introduced in Eq.~\eqref{eq:rain_gen_two} captures region-dependent rain streaks. 
In the real world, 
rain appearance is not only formed by individual rain streaks, but also by accumulating multiple rain streaks.
When rain accumulation is dense, the individual streaks
can not be seen clearly. This rain streak accumulation, whose visual effect is similar to
mist or fog, causes the atmospheric veiling effect as well as blur,
especially for distance scenes, as shown in Fig.~\ref{fig:real_rain_cases}.a.
Aside from rain accumulation, in many occasions, particularly
in heavy rain, rain streaks can have various shapes and
directions that overlap to each other, as shown in Fig.~\ref{fig:real_rain_cases}.a and~\ref{fig:real_rain_cases}.b.

\begin{figure}[tbp]
	\subfigure[]{	
		\centering\includegraphics[height=3.4cm]{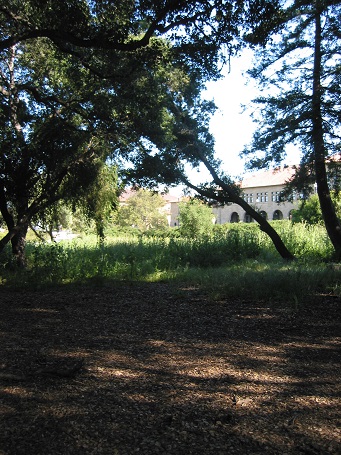}}	
	\subfigure[]{	
		\centering\includegraphics[height=3.4cm]{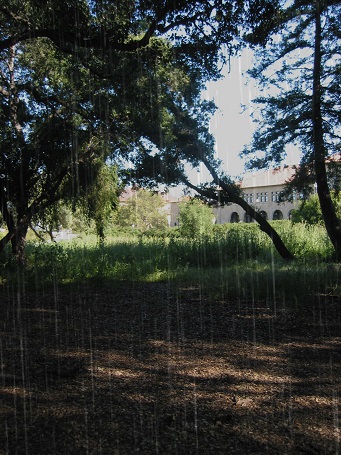}}
	\subfigure[]{	
		\centering\includegraphics[height=3.4cm]{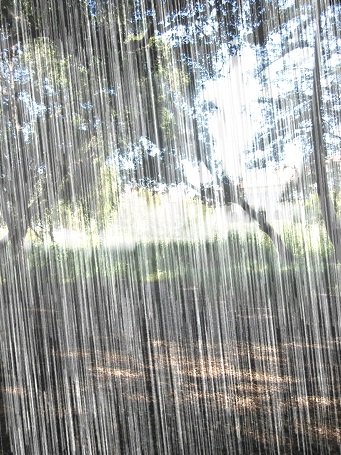}}
	\begin{small}
	\caption{Synthesized rain images (b) and (c) from (a) following the process of Eq.~\eqref{eq:rain_gen_two} and Eq.~\eqref{eq:rain_gen_multiple}, respectively. Instead of synthesizing the rain image with very sparse rain streaks in (b), our proposed rain image model considers both rain accumulation and multiple rain streaks overlapping in (c). Rain streaks with various shapes in different directions are overlapped in our synthesized result and distant scenes are invisible because of the rain accumulation.} 
	\label{fig:multiple_syn}
	\end{small}
	\vspace{-3mm}
\end{figure}

In this section, based on the model in Eq.~\eqref{eq:rain_gen1}, we create a model to accommodate rain streak accumulation:
\begin{equation}
	\label{eq:rain_gen_multiple}
	\mathbf{O} = \alpha\left(\mathbf{B} + \sum_{t=1}^{s}\widetilde{\mathbf{S}_t}\mathbf{R}\right)+\left(1-\alpha\right)\mathbf{A},
\end{equation}
where each $\widetilde{\mathbf{S}_t}$ is the direction- and
shape-consistent individually-visible rain streaks, $t$ is the
overlapping streak number, and $s$ is the number of shape and
direction consistent rain streaks within an image.
$\mathbf{A}$ is the global atmospheric light, $\alpha$ is the scene
transmission. Based on Eq.~\eqref{eq:rain_gen_multiple}, we can
synthesize the rain accumulation and heavy rain images as the training
data, which is closer to the appearance of natural rain, as shown in Fig.~\ref{fig:multiple_syn}.
Note that the atmospheric veiling effect is enforced on the integrated rain-contaminated image $\left(\mathbf{B} +
\sum_{t=1}^{s}\widetilde{\mathbf{S}_t}\mathbf{R}\right)$,
thus Eq.~\eqref{eq:rain_gen_multiple} implies that, we can handle rain accumulation and rain streak removal separately, which provides convenience for our training. 

In the following section, we first develop a deep convolutional network to detect and remove rain streaks from the rainy images having rain patterns explained by Eq.~\eqref{eq:rain_gen_multiple}. Then we consider the heavy rain situations and generalize the CNN  model to a recurrent one to perform iterative rain removal.

\begin{figure*}[t]	
		\centering\includegraphics[width=17cm]{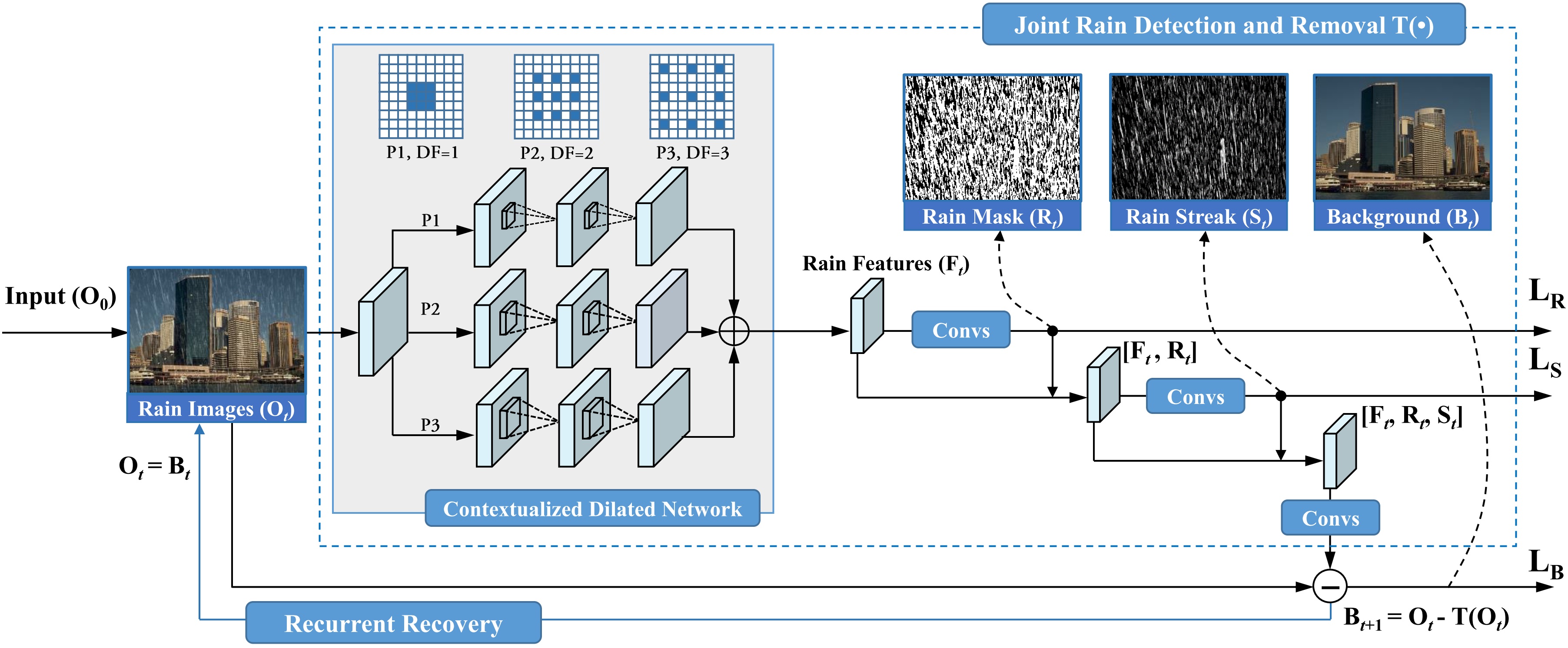}
	\caption{\C{The architecture of our proposed recurrent rain detection and removal. Each recurrence is a multi-task network to perform a joint rain detection and removal (in the blue dash box). In such a network, a contextualized dilated network (in the gray region) extracts rain features $\mathbf{F}_t$ from the input rain image $\mathbf{O}_t$. Then, $\mathbf{R}_t$, $\mathbf{S}_t$ and $\mathbf{B}_t$ are predicted to perform joint rain detection, estimation and removal. The contextualized dilated network has two features: 1) it takes a recurrent structure, which refines the extracted features progressively; 2) for each recurrence, the output features are aggregated from three convolution paths (P1, P2 and P3) with different dilated factors (DF) and receptive fields.}}
	\label{fig:flowchart}
	\vspace{-5mm}
\end{figure*}


\section{Convolutional Joint Rain Detection and Removal}
\label{sec:Multi_task}

We construct a convolutional multi-task network to perform \textbf{JO}int \textbf{R}ain
\textbf{DE}tection and \textbf{R}emoval~(JORDER) that solves the inverse problem in
Eq.~\eqref{eq:rain_gen_two} through end-to-end learning. Rain regions are first detected by JORDER
to further constrain the rain removal. To leverage more context without losing local details, we propose a novel network structure -- the contextualized dilated network -- for extracting the rain discriminative features and facilitating the following rain detection and removal. 

\subsection{Convolutional Multi-Task Networks for Joint Rain Detection and Removal}
\label{sec:mtl_jorder}

Relying on Eq.~\eqref{eq:rain_gen_two}, given the observed rain image
$\mathbf{O}$, our goal is to estimate $\mathbf{B}$, $\mathbf{S}$ and
$\mathbf{R}$.  Due to the nature of the ill-posed problem, it
leads to a maximum-a-posteriori (MAP) estimation:

\begin{equation}
	\label{eq:rain_map}
	\arg\min_{\mathbf{B},\mathbf{S},\mathbf{R}}||\mathbf{O} - \mathbf{B} - \mathbf{SR}||_2^2 + P_b (\mathbf{B})+ P_s (\mathbf{S})+ P_r (\mathbf{R}),
\end{equation}
where $P_b(\mathbf{B})$, $P_s (\mathbf{S})$ and $P_r (\mathbf{R})$ are the enforced priors on $\mathbf{B}, \mathbf{S}$ and $\mathbf{R}$,
respectively. Previous priors on $\mathbf{B}$ and $\mathbf{S}$ include
hand-crafted features, \textit{e.g.} cartoon texture
decomposition~\cite{ID}, and some data-driven models, such as sparse
dictionary~\cite{luo2015removing} and Gaussian mixture
models~\cite{li2016rain}. For deep learning methods, the priors about
$\mathbf{B}$, $\mathbf{S}$ and $\mathbf{R}$ are learned from the
training data and are embedded into the network implicitly. 

The estimation of $\mathbf{B},\mathbf{S}$ and $\mathbf{R}$ is
intrinsically correlated. Thus, the estimation of $\mathbf{B}$ benefits
from the predicted $\widehat{\mathbf{S}}$ and
$\widehat{\mathbf{R}}$. To address this, the natural choice is  to
employ a multi-task learning architecture, which can be trained using
multiple loss functions based on the ground truths of $\mathbf{R}$,
$\mathbf{S}$ and  $\mathbf{B}$ (see the blue dash box in Fig.~\ref{fig:flowchart} and please here ignore the subscript $t$ to denote the recurrence number).

As shown in the figure, we first exploit a contextualized dilated network to extract the rain feature
representation $\mathbf{F}$. Subsequently, $\mathbf{R}$, $\mathbf{S}$
and $\mathbf{B}$ are predicted in a sequential order, implying a
continuous process of rain streak detection, estimation
and removal. Each of them  is predicted
based on $\mathbf{F}$:
\begin{enumerate}
	\vspace{-2mm}
	\item $\mathbf{R}$ is estimated by two convolutions on $\mathbf{F}$,
	\vspace{-2mm}
	\item $\mathbf{S}$ is predicted by a convolution on the concatenation $\left[\mathbf{F},\widehat{\mathbf{R}}\right]$,
	\vspace{-2mm}
	\item $\mathbf{B}$ is computed from a convolution on the concatenation $\left[\mathbf{F},\widehat{\mathbf{R}}, \widehat{\mathbf{S}}, \mathbf{O}-\widehat{\mathbf{R}}\widehat{\mathbf{S}}\right]$. 
\end{enumerate}

There are several potential choices for the network
structures, such as estimating the three variables in the order of
$\mathbf{S}$, $\mathbf{R}$, $\mathbf{B}$, or in parallel (instead of
sequential). We compare some alternative architectures and demonstrate
the superiority of ours empirically in our supplementary material.

\subsection{Contextualized Dilated Networks}
For rain removal tasks, contextual information from an input image is
demonstrated to be useful for automatically
identifying and removing the rain streaks~\cite{2012_Huang_Context}. Thus, we propose a contextualized dilated network to aggregate context information at multiple scales for learning the rain features. 
The network gains contextual information in two ways: 1) it takes a recurrent structure, similar to the recurrent ResNet in~\cite{2016_Yang_DEGREE}, which provides an increasingly larger receptive field for the following layers; 2) in each recurrence, the output features aggregate representations from three convolution paths with different dilated factors and receptive fields.

Specifically, as shown in \C{the gray region of Fig.~\ref{fig:flowchart}}, the network first transforms the input rain image into feature space via the first convolution. Then, the network enhances the features progressively. In each recurrence, the results from the three convolution paths with different dilated factors are aggregated with the input features from the last recurrence via the identity forward.  The dilated convolution~\cite{YuKoltun2016} weights pixels with a step size of the dilated factor, and thus increases its receptive field without losing resolution. Our three dilated paths consist of two convolutions with the same kernel size $3\times3$. However, with different dilated factors, different paths have their own receptive field. As shown in the~\C{top part of the gray region in Fig.~\ref{fig:flowchart}, path P$_2$ consists of two convolutions with the dilated factor $2$. The convolution kernel is shown as the case of DF$=2$. Thus, cascading two convolutions, the three paths have their receptive fields of $5\times5$, $9\times9$ and $13\times13$.}

\subsection{Network Training}
Let $\mathbf{F}_{\text{rr}}(\cdot),\mathbf{F}_{\text{rs}}(\cdot)$ and
$\mathbf{F}_{\text{bg}}(\cdot)$ denote the inverse recovery functions
modeled by the learned network to generate the estimated rain streak
binary map~$\widehat{\mathbf{R}}$, rain streak
map~$\widehat{\mathbf{S}}$ and background image $\widehat{\mathbf{B}}$
based on the input rain image $\mathbf{O}$. We use $\mathbf{\Theta}$
to collectively represent all the parameters of the network.

We use $n$ sets of corresponding rain images, background images, rain
region maps and rain streak maps $\left\{\left(\mathbf{o}_i,
\mathbf{g}_i, \mathbf{r}_i, \mathbf{s}_i\right) \right\}_{i=1}^{n}$
for training.  
We adopt the following joint loss function to train the network parametrized
by $\mathbf{\Theta}$ such that it is capable to jointly estimate
$\mathbf{r}_i$, $\mathbf{s}_i$ and $\mathbf{g}_i$ based on rain image
$\mathbf{o}_i$: 
\begin{align}
	\label{eq:loss}
	L(\mathbf{\Theta})=&\frac{1}{n}\sum_{i=1}^{n}\left(||\mathbf{F}_{\text{rs}}\left(\mathbf{o}_{i};\mathbf{\Theta}\right)-\mathbf{s}_i||^2 + \lambda_1 ||\mathbf{F}_{\text{bg}}\left(\mathbf{o}_{i};\mathbf{\Theta}\right)-\mathbf{g}_i||^2\right. \nonumber \\
	& \left.-\lambda_2 \left(\log\widehat{\mathbf{r}}_{i,1}\mathbf{r}_{i,1}+\log(1-\widehat{\mathbf{r}}_{i,2})(1-\mathbf{r}_{i,2})\right)\right), \nonumber \\
	& \text{with } \widehat{\mathbf{r}}_{i,j} =  \frac{\exp{\left\{\mathbf{F}_{\text{rs},j}\left(\mathbf{o}_{i};\mathbf{\Theta}\right)\right\}}}{\sum_{k=1}^{2}\exp{\left\{\mathbf{F}_{\text{rs},k}\left(\mathbf{o}_{i};\mathbf{\Theta}\right)\right\}}}, j\in\left\{1,2\right\}.
\end{align}
Here $\lambda_1$ and $\lambda_2$ are the parameters to
balance the importance among estimating $\mathbf{S}$, $\mathbf{R}$ and
$\mathbf{B}$. The network is trained to minimize the above loss, via error back-propagation. 

\section{Removing Rains from Real Rain Images}
\label{sec:IEFN}

In the previous section, we construct a convolutional multi-task learning network to jointly detect and remove rain streaks from rain images. 
In this section, we further enhance our network to
handle rain accumulation and rain streaks that possibly have various
shapes and directions in one image. 

\subsection{Recurrent Rain Detection and Removal}
\label{sec:IEF}

The recurrent JORDAR model can be understood as a cascade of the
convolutional joint rain detection and removal networks to perform 
progressive rain detection and removal and recover the image with increasingly better visibility. 

\paragraph{Architecture.} 
We define the process of the network in the blue dash box of
Fig.~\ref{fig:flowchart} that generates the residual image
between $\mathbf{O}$ and $\mathbf{B}$ as
$\mathbf{T}\left(\cdot\right)$. Then,  the recurrent rain detection and removal works as follows,
\begin{eqnarray}
	\label{eq:formulation_IEF}	
	\left[\mathbf{\epsilon}_t, \mathbf{R}_t, \mathbf{S}_t\right] &=& \mathbf{T}\left(\mathbf{O}_{t}\right),\nonumber \\
	\mathbf{B}_{t} &=& \mathbf{O}_{t} - \mathbf{\epsilon}_t, \\
	\mathbf{O}_{t+1} &=& \mathbf{B}_{t}. \nonumber
\end{eqnarray}
In each iteration $t$, the predicted residue $\mathbf{\epsilon}_t$ is
accumulated and propagated to the final estimation via updating
$\mathbf{O}_{t}$ and $\mathbf{B}_{t}$. Note that, although the estimated rain mask $\mathbf{R}_t$ and streak $\mathbf{S}_t$ are not casted into the next recurrence directly. However, the losses to regularize them in fact provide strong side information for learning an effective $\mathbf{T}\left(\cdot\right)$. The final estimation can be
expressed as:
\begin{eqnarray}
	\mathbf{B}_{\tau} = \mathbf{O}_{0} + \sum_{t=1}^\tau\mathbf{\epsilon}_t,
\end{eqnarray}
where $\tau$ is the total iteration number. Hence, the process removes the rain streak progressively, part by part, based on the intermediate results from the  previous step. The complexity of rain removal in each iteration is consequently reduced, enabling better estimation, especially in the case of heavy rains. 

\paragraph{Network Training.} 
The recurrent JORDAR network introduces an extra time variable $t$ 
to the loss function $L(\mathbf{\Theta})$ in Eq.~\eqref{eq:loss} and gives $L(\Theta_t, t)$,
where $L(\mathbf{\Theta}_0,0) = L(\mathbf{\Theta}_0)$. When $t>1$, $L(\mathbf{\Theta}_t,t)$ is equivalent to $L(\mathbf{\Theta})$ that replaces $\mathbf{o}_i$ and
$\mathbf{\Theta}$ by $\mathbf{o}_{i,t}$ and $\mathbf{\Theta}_t$,
respectively, where $\mathbf{o}_{i,t}$ is generated from the $t$-th
iterations of the process Eq.~\eqref{eq:formulation_IEF} on the initial $\mathbf{o}_i$. Then, the total loss function $L_\text{Iter}$ for training $\mathbf{T}$ is

\begin{equation}
	L_\text{Iter}(\left\{\mathbf{\Theta}_0, ..., \mathbf{\Theta}_\tau \right\}) = \sum_{t=0}^{\tau}L(\mathbf{\Theta}_t, t).
\end{equation}

\subsection{Joint Derain and Dehaze}
\label{sec:alter_dehaze}

Distant rain streaks accumulate and form an atmospheric veil, 
causing visibility degradation. To resolve this,  clearing up the atmospheric veil, which is similar to dehazing or defogging, is necessary.  

Eq.~\eqref{eq:rain_gen_multiple} suggests that dehazing should be the first step in the process of joint deraining and dehazing. Thus, we propose to estimate $\left(\mathbf{B} + \sum_{t=1}^{s}\widetilde{\mathbf{S}_t}\right)$
first. However,  placing dehazing as a preprocessing has complicated
effects on deraining, since all rain streaks (including the ones that
are already sharp and clearly visible) are boosted, making  the 
streaks look different from those in the training images. Hence, in
our proposed pipeline, we derain first, then followed by dehazing and at last finished with deraining. This, as it turns
out, is beneficial, since dehazing will make the appearance of less
obvious rain streaks (which are likely unnoticed by the first round of
deraining) become more obvious. 

We implement a dehazing network based on the structure of contextualized dilated network, with only one recurrence, trained with the
synthesized data generated with the random background reliance and
transmission value~\cite{DehazeNet}. We find that the sequential process of 
derain-dehaze-derain is generally effective (see the supplementary material for the evaluation of other possible sequences). The reason is that some obvious rain
streaks, noises and artifacts are removed in the first round
deraining. Then, the dehazing cleans up the rain accumulation,
enhances the contrast and visibility, and at the same time boosts weak
rain streaks. The subsequent deraining removes these boosted
rain streaks, as well as artifacts caused by dehazing, making the
results cleaner.  

\section{Experimental Results}
\label{sec:Exp}

\begin{figure}[tbp]
	\centering
	\subfigure[Synthesized streak following~\cite{garg2006photorealistic}]{
		\includegraphics[height=2.6cm]{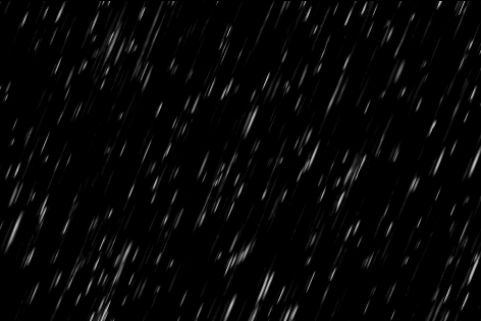}}
	\subfigure[Synthesized sharp line streak]{
		\includegraphics[height=2.6cm]{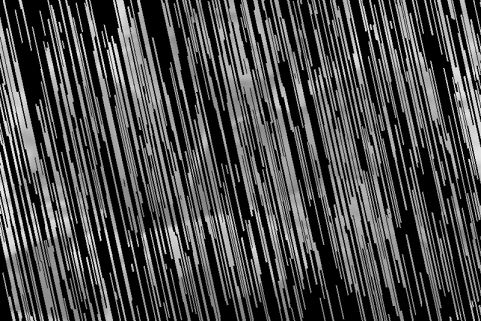}}
	\\
	\subfigure[An example from \textit{Rain100L}]{
		\includegraphics[height=2.6cm]{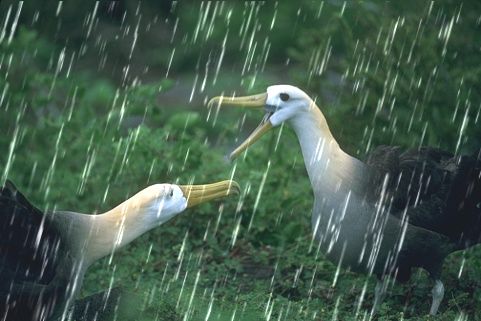}}
	\subfigure[An example from \textit{Rain100H}]{
		\includegraphics[height=2.6cm]{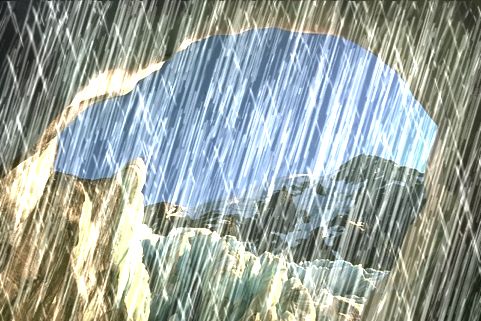}}
	\caption{The examples of synthesized rain streaks and rain images.}
	\label{fig:dataset_examples}
	\vspace{-4mm}
	
\end{figure}

\paragraph{Datasets.} 
We compare our method with state-of-the-art methods on a few benchmark
datasets: (1)
\textit{Rain12}~\footnote{http://yu-li.github.io/}~\cite{li2016rain}, which includes 12 synthesized rain images with only one type of rain streak;~\textit{Rain100L}, which is the synthesized data set with only one type of rain streak (Fig.~\ref{fig:dataset_examples}.c); 
(2)  \textit{Rain20L}, which is a subset of Rain100L used for testing the potential network structures in the supplementary material; 
(3) \textit{Rain100H},  which is our synthesized data set
with five streak directions (Fig.~\ref{fig:dataset_examples}.d).
Note, while it is rare for a real rain image to contain rain streaks
in many different directions, synthesizing this kind of images for training is observed to boost the capacity of the network.  

The images for synthesizing \textit{Rain100L}, \textit{Rain20L}
and \textit{Rain100H} are selected from BSD200~\cite{MartinFTM01}. The
dataset for training our network and another deep learning baseline
-- SRCNN for deraining -- is BSD300, excluding the
ones appeared in \textit{Rain12}. The rain streaks are synthesized in
two ways: (1) the photorealistic rendering techniques proposed
by~\cite{garg2006photorealistic} as shown in
Fig.~\ref{fig:dataset_examples}.a; (2) the simulated sharp line
streaks along a certain direction with a small variation within an
image as shown in Fig.~\ref{fig:dataset_examples}.b. We will release our training and testing sets, as well as their synthesis codes for public in the future. 

\vspace{-4mm}

\def\height{3.5cm}
\def\name{p23}
\def\namet{p19}
\def\nameth{p11}

\begin{figure*}[!tbp]
	\begin{center}
		{ 	\subfigure{
				\includegraphics[height=\height,clip]{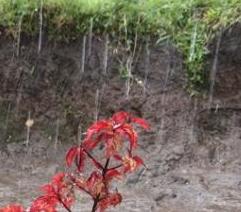}}
			\subfigure{
				\includegraphics[height=\height,clip]{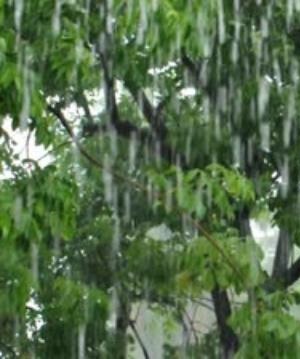}}
			\subfigure{
				\includegraphics[height=\height,clip]{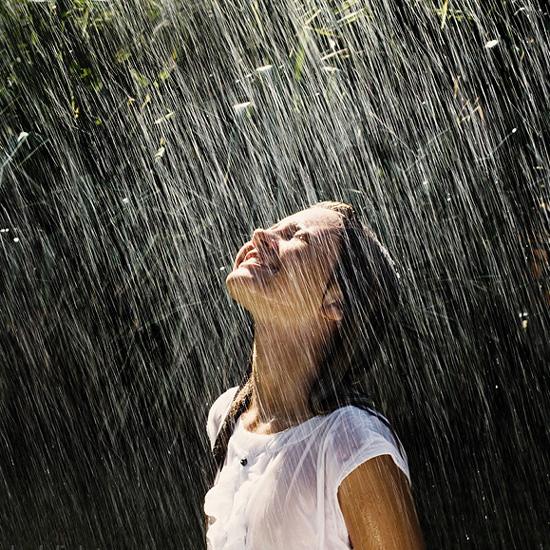}}			
			\subfigure{
				\includegraphics[height=\height,clip]{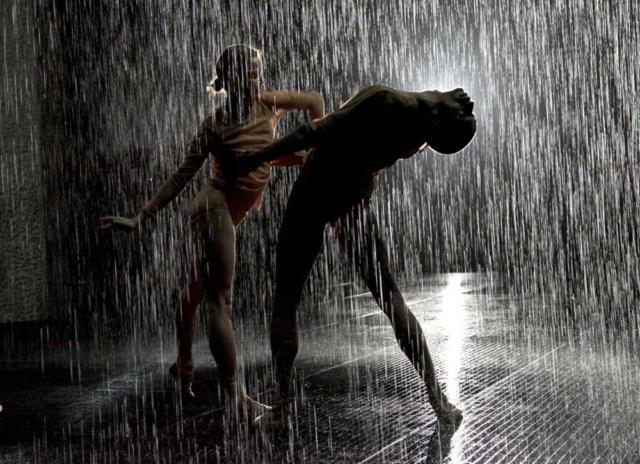}}					
			\\	
			\vspace{-3mm}
			\subfigure{
				\includegraphics[height=\height,clip]{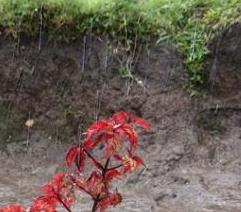}}
			\subfigure{
				\includegraphics[height=\height,clip]{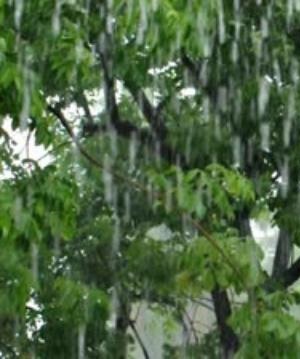}}
			\subfigure{
				\includegraphics[height=\height,clip]{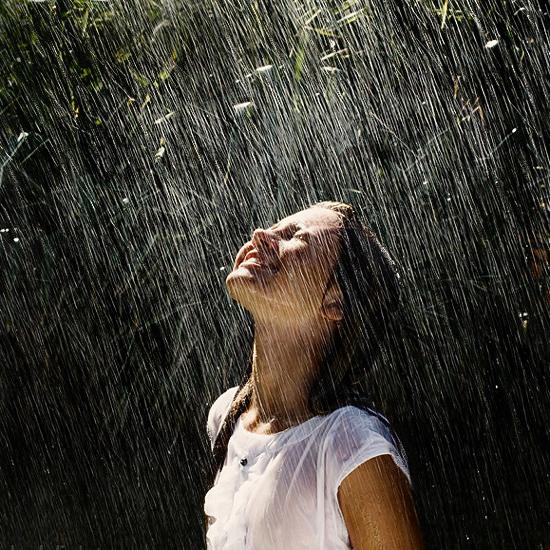}}			
			\subfigure{
				\includegraphics[height=\height,clip]{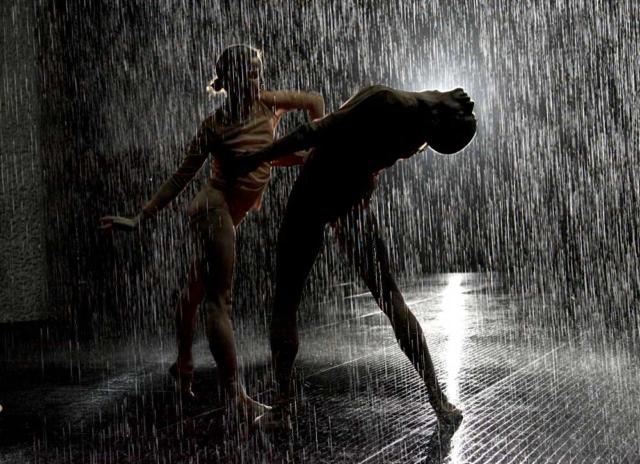}}     
			\\ 			
			\vspace{-3mm}
			\subfigure{
				\includegraphics[height=\height,clip]{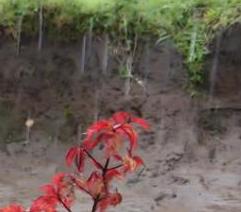}}
			\subfigure{
				\includegraphics[height=\height,clip]{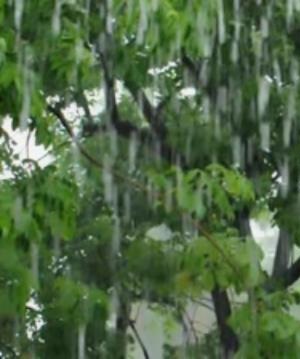}}
			\subfigure{
				\includegraphics[height=\height,clip]{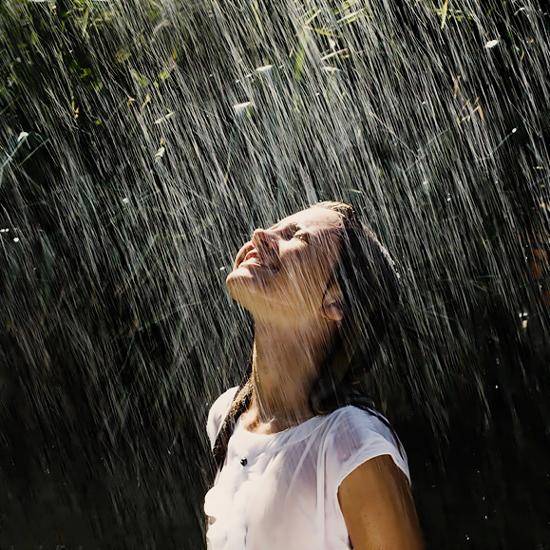}}				
			\subfigure{
				\includegraphics[height=\height,clip]{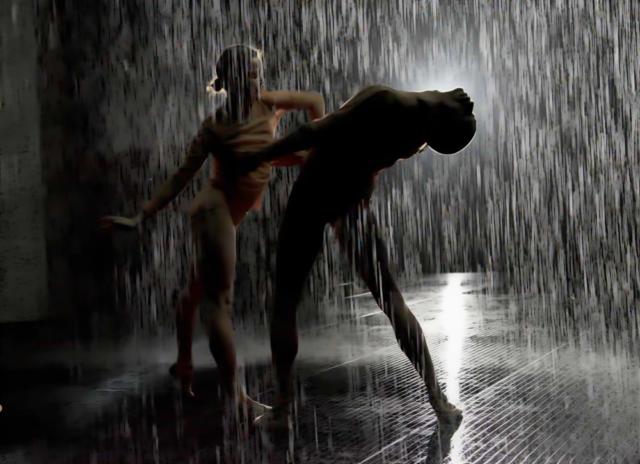}}			
			\\
			\vspace{-3mm}			
			\subfigure{
				\includegraphics[height=\height,clip]{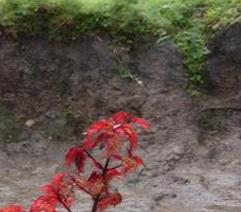}}
			\subfigure{
				\includegraphics[height=\height,clip]{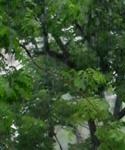}}
			\subfigure{
				\includegraphics[height=\height,clip]{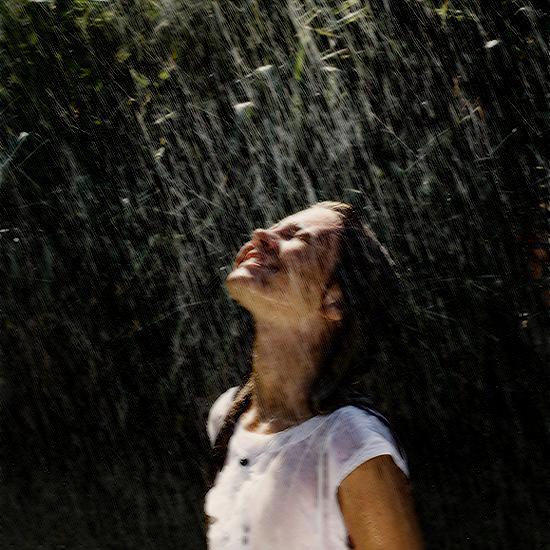}}			
			\subfigure{
				\includegraphics[height=\height,clip]{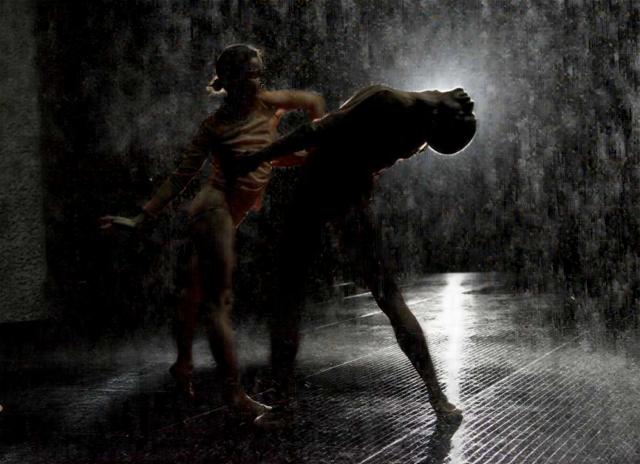}}			
		} 
	\end{center}
	\vspace{-3mm}
	
	\caption{Results of different methods on real images. From top to down: rain image, DSC, LP and JORDER-R.}
	\label{fig:practical}
	\vspace{-3mm}
\end{figure*}

\paragraph{Baseline Methods.}
We compare the four versions of our approaches, JORDER-~(one version of our methods that has only one convolution path in each recurrence without using dilated convolutions), JORDER
(Section~\ref{sec:Multi_task}), JORDER-R (Section~\ref{sec:IEF}),
JORDER-R-DEHAZE (Section~\ref{sec:alter_dehaze}) with five
state-of-the-art methods: image decomposition~(ID)~\cite{ID},
CNN-based rain drop removal~(CNN) \cite{Eigen_2013_ICCV},
discriminative sparse coding~(DSC)~\cite{luo2015removing}, layer
priors~(LP)~\cite{li2016rain} and a common CNN baseline for image
processing -- SRCNN~\cite{ACMMM_Caffe_2014}, trained for deraining.  
SRCNN is implemented and trained by ourselves, while other methods are
kindly provided by the authors. 

For the experiments on synthesized data, two metrics Peak
Signal-to-Noise Ratio~(PSNR)~\cite{huynh2008scope} and Structure
Similarity Index~(SSIM)~\cite{wang2004image} are used as comparison
criteria. We evaluate the results only in the luminance channel, which
has a significant impact on the human visual system to perceive the
image quality.  

\paragraph{Quantitative Evaluation.}

Table~\ref{tab:psnr_ssim_result} shows the results
of different methods on \textit{Rain12}. As observed, our method
considerably outperforms other methods in terms of both PSNR and SSIM. 
Table~\ref{tab:hard_case} presents the results of JORDER and
JORDER-R on \textit{Rain100H}. 
Note that, our JODDER-R is designed to handle such hard cases, thus achieves considerably better results than other compared methods. The PSNR of JORDER-R gains over JORDER more than 1dB. Such a large gain demonstrates that the recurrent rain detection and removal significantly boosts the performance on synthesized heavy rain images.

\begin{table}[tbp]
	\small
	\centering
	\caption{PSNR and SSIM results among different rain streak removal methods on \textit{Rain12} and \textit{Rain100L}.}
	\vspace{3mm}
	
	\begin{tabular}{c|cc|cc}
		\hline
		Baseline & \multicolumn{2}{c|}{\textit{Rain12}} & \multicolumn{2}{c}{\textit{Rain100L}} \\
		\hline
		Metric & PSNR  & SSIM  & PSNR  & SSIM \\
		\hline
		ID    & 27.21 & 0.75  & 23.13 & 0.70  \\
		DSC   & 30.02 & 0.87  & 24.16 & 0.87  \\
		LP    & 32.02 & 0.91  & 29.11 & 0.88  \\
		CNN   & 26.65 & 0.78  & 23.70 & 0.81  \\
		SRCNN & 34.41 & 0.94  & 32.63 & 0.94  \\
		\hline
		JORDER- & 35.86 & 0.95 & 35.41 & 0.96  \\
		JORDER & \textbf{36.02} & \textbf{0.96} & \textbf{36.11} & \textbf{0.97} \\
		\hline
	\end{tabular}%
	\label{tab:psnr_ssim_result}%
	\vspace{-6mm}
\end{table}%

\begin{table}[tbp]
	\small
	\centering
	\caption{PSNR and SSIM results among different rain streak removal methods on \textit{Rain100H}.}
	\vspace{3mm}	
	
	\begin{tabular}{c|ccc}
		\hline
		Metric & ID & LP & DSC \\
		\hline
		PSNR  & 14.02  &  14.26   &  15.66  \\ 
		SSIM  & 0.5239 &  0.4225  &  0.5444   \\ 
		\hline
		Metric & JORDER- & JORDER & JORDER-R \\
		\hline
		PSNR  & 20.79  & 22.15 & \textbf{23.45} \\
		SSIM  &  0.5978 & 0.6736 & \textbf{0.7490} \\
		\hline		
	\end{tabular}%
	\label{tab:hard_case}%
	\vspace{-4mm}	
\end{table}%

\begin{figure*}[!tbp]
	\centering
	\subfigure{
		\includegraphics[height=3.5cm]{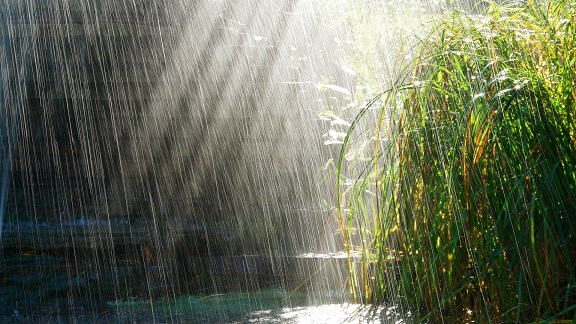}}
	\subfigure{
		\includegraphics[height=3.5cm]{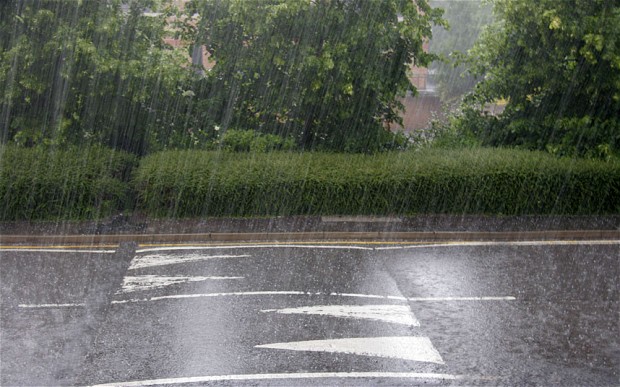}}		
	\subfigure{
		\includegraphics[height=3.5cm]{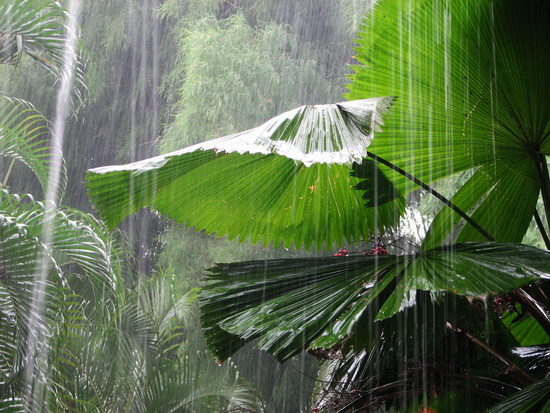}}
	\\
	\subfigure{
		\includegraphics[height=3.5cm]{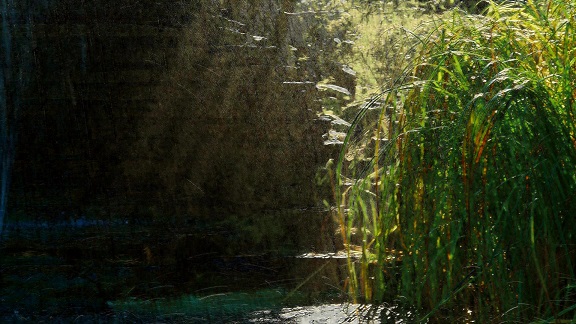}}
	\subfigure{
		\includegraphics[height=3.5cm]{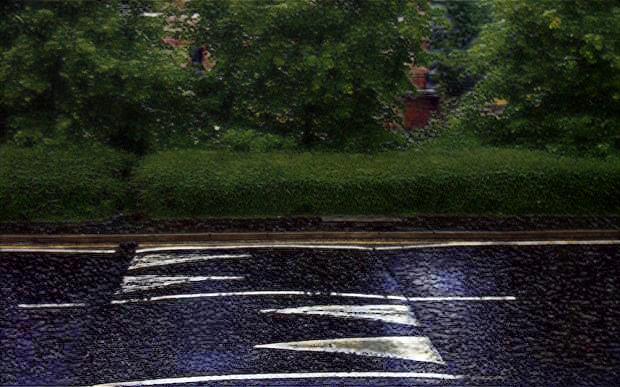}}			
	\subfigure{
		\includegraphics[height=3.5cm]{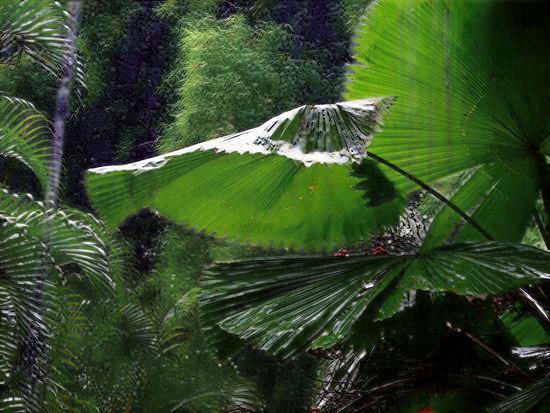}}
	\caption{The examples of JORDER-R-DEHAZE on heavy rain (left two images) and mist images (right two images).}
	\label{fig:extreme2}
\end{figure*}

%


\paragraph{Qualitative Evaluation.}
Fig.~\ref{fig:practical} shows the results of real images. For
fair comparisons, we use JORDER-R to process these rain images and
do not handle the atmospheric veils on these results, to
be consistent with other methods. As observed, our method
significantly outperforms them and is successful in removing the
majority of rain streaks.  

We also compare all the methods in two extreme cases: dense rain
accumulation, and  heavy rain as shown in Fig.~\ref{fig:extreme2}. Our method achieves promising results in
removing the majority of rain streaks, enhancing the visibility and
preserving details. All the results and related codes will be publicly available in the future.

Table~\ref{tab:runing_time} compares the running time of several state-of-the-art methods. All baseline methods are implemented in MATLAB. Our methods are implemented on the Caffe's Matlab wrapper.  CNN rain drop and some versions of our methods are implemented on GPU, while others are
based on CPU. Our GPU versions is computationally efficient. The CPU version of JORDER, a lightest
version of our method, takes up the shortest running time among all
CPU-based approaches. In general, our methods in GPU are capable of dealing
with a $500\times500$ rain image less than 10s, which is considerably faster than the existing methods.

\begin{table}[!tbp]
	\small
	\centering
	\caption{The time complexity~(in seconds) of JORDER compared with state-of-the-art methods. JR and JRD denote JORDER-R and JORDER-R-DEHAZE, respectively. (G) and (D) denote the implementation on GPU and GPU, respectively.}
	\vspace{3mm}
		
	\begin{tabular}{c|cccc}
		\hline
		Scale & CNN (G) & ID    & DSC   & LP  \\
		\hline
		80*80 & 0.85  & 449.94  & 14.32  & 35.97 \\    
		500*500 & 6.39  & 1529.85  & 611.91  & 2708.20 \\
		\hline
		Scale & JORDER (C) & JORDER (G) &  JR (G) & JRD (G) \\
		\hline
		80*80 & 2.97 & 0.11 & 0.32 & 0.72 \\
		500*500 & 69.79 & 1.46 & 3.08 & 7.16 \\
		\hline
	\end{tabular}%
	\label{tab:runing_time}%
	\vspace{-4mm}
	
\end{table}%

\paragraph{Evaluation on joint derain and dehaze.}

\begin{figure}[!tbp]
	\centering
	\subfigure[Rain image]{
		\includegraphics[height=1.7cm]{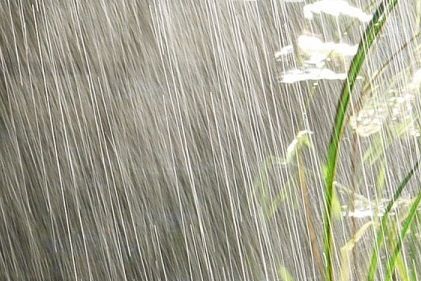}}
	\subfigure[Derain]{
		\includegraphics[height=1.7cm]{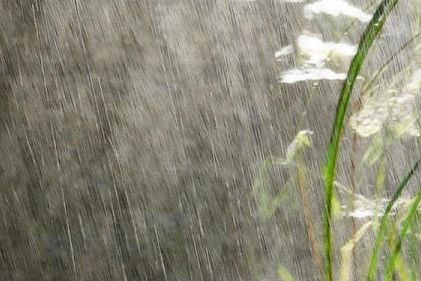}}
	\subfigure[Derain-Derain]{
		\includegraphics[height=1.7cm]{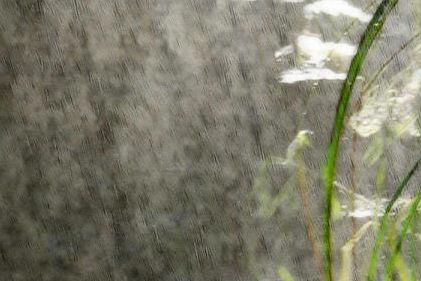}}		
	\\
	\vspace{-3mm}
	\subfigure[Derain-dehaze]{
		\includegraphics[height=1.7cm]{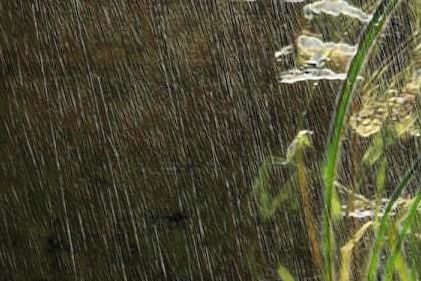}}
	\subfigure[Dehaze-derain]{
		\includegraphics[height=1.7cm]{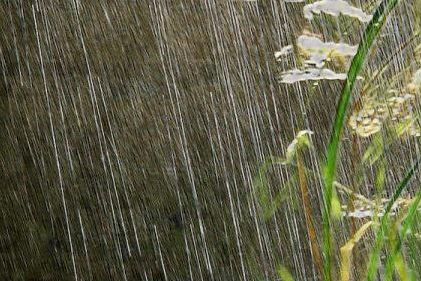}}
	\subfigure[Derain-dehaze-derain]{
		\includegraphics[height=1.7cm]{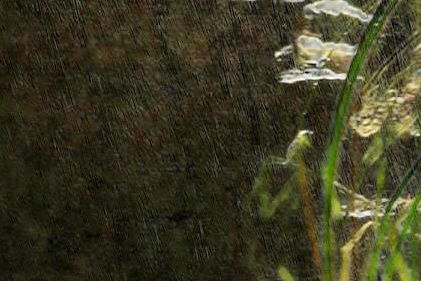}}		
	\caption{The results of JORDER-R-DEHAZE in different orders.}
	\label{fig:order}
	\vspace{-6mm}
	
\end{figure}

Fig.~\ref{fig:order} shows the significant superiority of our method~(f), in an order of derain-dehaze-derain, than other potential combinations~((b)-(e)).

\section{Conclusion and Future Works}
\label{sec:conclusion}

In this paper, we have introduced a new deep learning based method to
effectively learn to joint remove rain from a single image, even in the presence of
rain streak accumulation and  heavy rain. A new region-dependent
rain image model is proposed for additional rain detection
and  is further extended to simulate rain accumulation and heavy
rains. Based on this model, we developed a fully convolutional
network that jointly detect and remove rain. Rain regions
are first detected by the network which naturally provides additional
information for rain removal. To restore images captured in the
environment with both rain accumulation and heavy rain, we introduced an
recurrent rain detection and removal network that progressively removes rain
streaks, embedded with a dehazing network to remove atmospheric
veils. Evaluations on real images demonstrated that our method outperforms
state-of-the-art methods significantly.

\bibliographystyle{abbrv}
\bibliography{sigproc}
\end{document}